\documentclass[runningheads]{llncs}
\usepackage{subfig}
\usepackage{graphicx}
\usepackage{amsmath}
\usepackage{amsfonts}
\usepackage[utf8]{inputenc}
\usepackage[greek.polutoniko,english]{babel}

\begin{document}
\title{A Robust Deep Ensemble Classifier for Figurative Language Detection}
\author{Potamias Rolandos-Alexandros \and
Siolas Georgios \and
Stafylopatis Andreas}

\authorrunning{R. Potamias et al.}

%
\institute{
Artificial Intelligence and Learning Systems Laboratory,\\
School of Electrical and Computer Engineering,\\
National Technical University of Athens,  Greece\\
\email{rolpotamias@gmail.com, 	gsiolas@islab.ntua.gr, andreas@cs.ntua.gr}\\
}
\maketitle              
\begin{abstract}
Recognition and classification of Figurative Language (FL) is an open problem of Sentiment Analysis in the broader field of Natural Language Processing (NLP) due to the contradictory meaning contained in phrases with metaphorical content. The problem itself contains three interrelated FL recognition tasks: sarcasm, irony and metaphor which, in the present paper, are dealt with advanced Deep Learning (DL) techniques. First, we introduce a data prepossessing framework towards efficient data representation formats so that to optimize the respective inputs to the DL models. In addition, special features are extracted in order to characterize the syntactic, expressive, emotional and temper content reflected in the respective social media text references. These features aim to capture aspects of the social network user’s writing method. Finally, features are fed to a robust, Deep Ensemble Soft Classifier (DESC)  which is based on the combination of different DL techniques. Using three different benchmark datasets (one of them containing various FL forms) we conclude that the DESC model achieves a very good performance, worthy of comparison with relevant methodologies and state-of-the-art technologies in the challenging field of FL recognition.

\end{abstract}
%
%
%
\section{Introduction}
Figurative language (FL), as a linguistic phenomenon, refers to the contradiction between the literal and non-literal meaning of a sentence. Detection of FL is an open problem of computational linguistics mostly because of the difference or distance between the form and the actual meaning of a sentence. In the present paper, we address sentiment analysis of FL as well as tracking three most common types of FL, irony (Ancient Greek: \textgreek{εἰρωνεία}), sarcasm (\textgreek{σαρκασμός}) and metaphor (\textgreek{μεταφορά}). 
Bellow are three indicative FL examples: 
\begin{itemize}
\item\textit{So many useless classes , great to be student (sarcasm) }
\item\textit{You're about as genuine as a used car salesman. (metaphor) }
\end{itemize}
There are two types of irony, situational and verbal, depending on its use. Verbal irony is defined as the FL phenomenon in which the author denotes exactly the opposite of what he/she means. Aristotle himself described irony as an ``refined insult'' highlighting its trenchant use. In contrast with irony, sarcasm cannot be easily defined. Both Oxford\footnote{\url{https://en.oxforddictionaries.com/definition/sarcasm}} and Merriam Webster\footnote{\url{ https://www.merriam-webster.com/dictionary/sarcasm}} Dictionaries do not dissever sarcasm from irony, describing sarcasm as ``the use of irony to mock or convey contempt". Sarcastic comments tend to be more condescending in contrast with ironic ones that mostly denote humour. On the other hand, metaphor, a super set of irony and sarcasm, is a figure of speech that when taken in its literal sense makes no or little meaning, with its underlying meaning to be still easily understood. These definitions of FL forms imply the ambiguity separating literal and non-literal comments. 

Despite that all forms of FL are well studied linguistic phenomena \cite{w._gibbs_psycholinguistics_1986}, computational approaches fail to identify the polarity of them within a text. The influence of FL in sentiment classification emerged on SemEval-2014 Sentiment Analysis task \cite{rosenthal_semeval-2014_2014}. Results show that Natural Language Processing (NLP) systems effective in most other tasks see their performance drop when dealing with figurative forms of language. Thus, methods capable of detecting, separating and classifying forms of FL would be valuable building blocks for a system that could ultimately provide a full-spectrum sentiment analysis of natural language.    

In this paper we present a new method for FL detection. The novelty of our approach is founded on: (a) the integration of different deep leaning architectures (LSTMs, DNN); (b) the introduction of an ensemble between different neural classifiers; and (c) the combination of different representations including word-embeddings and engineered features.

\section{Literature Review}
Despite all forms of FL have been studied independently by the Machine Learning community, none of the proposed systems have been tested on more than one problem. Related work on the impact of FL in sentence sentiment classification problems are usually categorized with respect to their subject: irony, sarcasm detection and sentiment analysis of figurative language. Many researchers tend to treat sarcasm and irony as an identical phenomenon in their works but we will investigate each subject separately.   
\subsection{Irony and Sarcasm Detection} Irony detection was first explored by Reyes  \cite{reyes_humor_2012,reyes_multidimensional_2013}. Indeed, his work carried out the first comprehensive studies on FL from a computer science perspective. He approached irony as a linguistic phenomenon, undistinguished from sarcasm, that contains something unexpected or contradictory. Decision trees were used to classify features indicating unexpectedness like emotional words, contradictory terms and punctuation. A similar process was followed by Barbieri \cite{barbieri_modelling_2014}, where he performed sarcasm detection in topics such as politics, education and humor. Data was collected from Twitter using the hashtags  \#sarcasm, \#politics, \#education, \#humour. He calculates a measure of the unexpectedness and possible ambiguities, based on words that are mainly used in spoken language based on the American National Corpus Frequency Data\footnote{http://www.anc.org/data/anc-second-release/frequency-data/} as well as the tweets morphology. Data classification is performed using Random Forests and Decision Trees. Buschmeir \cite{buschmeier_impact_2014} used Logistic Regression to recognize irony and focused on the unexpectedness factor which is considered as an emotional imbalance between words in the text. In another direction, an attentive deep learning approach was implemented by Huang \cite{huang_irony_2017} using Google's pre-trained Word2vec Embeddings \cite{mikolov2013linguistic}. A pattern-based method to detect irony was proposed by Carvalho \cite{carvalho_clues_2009}, using n-grams and combinations of adverbs and acronyms that indicate humor, as features. A context approach was proposed by Wallace \cite{wallace_sparse_2015}, to indicate irony on Reddit posts using previous and following comments. On \textit{Semantic Evaluation Workshop-2018 task `Irony Detection in English Tweets'} \cite{hee_semeval-2018_2018} four teams showcased remarkable results. Team named \textit{Thu-Ngn} used a fully-connected Long short-term memory (LSTM) with pretrained Word Embeddings and a combination of syntactic and emotional features. An unweighted average between character and word level bidirectional LSTMs was proposed by the Ntua-Slp team, using an attention layer to determine irony. A majority ensemble classifier consisting of Linear Regression and SVM was proposed by WLV, using Word-Emoji embeddings as well as features highlighting the tweet's emotional background. Finally, the NLPRL-IITBHU team proposed features based on the opposition and disharmony of words. Ghosh \cite{ghosh_sarcastic_2015} claims that the key for sarcasm detection is the contradiction between words within the same tweet. He implements a SVM kernel function, based on similarity measure of sarcastic tweets as well as word2vec embeddings. Davidov's  \cite{davidov_semi-supervised_2010} semi-supervised pattern-based approach used attributes like words frequency and content words to classify them using k-Nearest Neighbours. On the other hand,  Ibanez \cite{gonzalez-ibanez_identifying_2011} proposed a different procedure considering sarcasm as a binary classification task, against positive-negative comments. Features are both lexical, extracted using WordNet-Affect and LWIC, and so-called ``pragmatic factors'', reflecting users sentiment. A behavioral approach was implemented by Rajadesingan \cite{rajadesingan_sarcasm_2015}, recording user's background information. In this work both behavioral and sentiment contrast features fed an SVM classifier. Word Embeddings combined with (Convolutional Neural Networks) CNN-LSTM units were used by Kumar\cite{kumar_having_2017} and Ghosh\&Veale \cite{ghosh_fracking_2016} resulting state-of-the-art performance. 
\subsection{Sentiment Analysis on Figurative Language}
The Semantic Evaluation Workshop-2015 \cite{ghosh_semeval-2015_2015} proposed a joint task to evaluate the impact of FL in sentiment analysis on ironic, sarcastic and metaphorical data from tweets. The ClaC team exploited four lexicons to extract attributes as well as syntactic features to identify sentiment polarity. The UPF team used regression to classify features extracted using lexicons such as SentiWordNet and DepecheMood. The LLT-PolyU team used semi-supervised Regression and Decision Trees, given uni-gram and bi-gram models. Possible word contradiction at short distances was taken into account. A SVM-based classifier was used by the Elirf team, given n-gram and Tf-idf features. In addition, lexicons such as Affin, Pattern and Jeffrey10 were also utilized. Finally, the LT3 team used an ensemble Regression and SVM semi-supervised classifier. The features consisted mainly of a range of lexical data combined with WordNet and DBpedia11.

\section{A Deep Learning Architecture for Figurative Language Recognition}
Our work proposes an ensemble architecture consisting of three deep models detailed in the present section, a BiLSTM, an AttentionLSTM and a Dense NN. Both BiLSTM and AttentionLSTM are sequential architectures comprising  LSTM cells, fed with pre-trained GloVe Word Embeddings \cite{pennington_glove:_2014}. The DNN model is feeded with several features extracted from each tweet, such as uni-grams, bi-grams Tf-Idf representations of text combined with syntactic, demonstrative, sentiment, mood and readability features to track FL. The proposed robust ensemble classifier (DESC) uses soft classification techniques. 
\subsection{BiLSTM}
\begin{figure}
\centering
\begin{minipage}{.48\textwidth}
  \centering
    \includegraphics[scale=0.12, angle=-90]{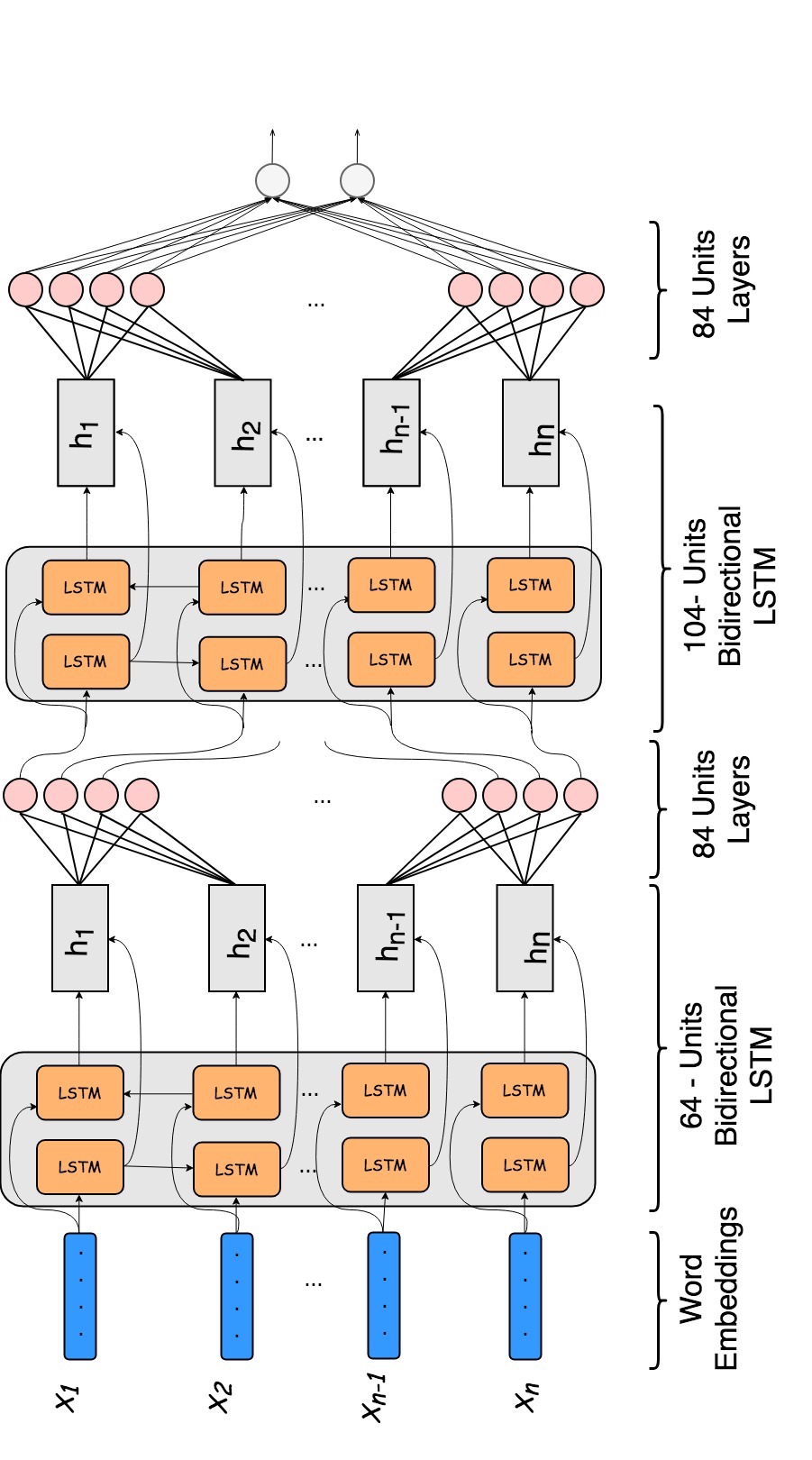}
    \vspace{0.36cm}
  \caption{ BiLSTM architecture consisting of two bidirectional LSTM layers, LeakyReLU activated and fully connected with a dense layer.} 
  \label{fig:bilstm}
\end{minipage}%
\hspace{0.15cm}
\begin{minipage}{.47\textwidth}
  \centering
      \includegraphics[scale=0.13, angle=-90  ]{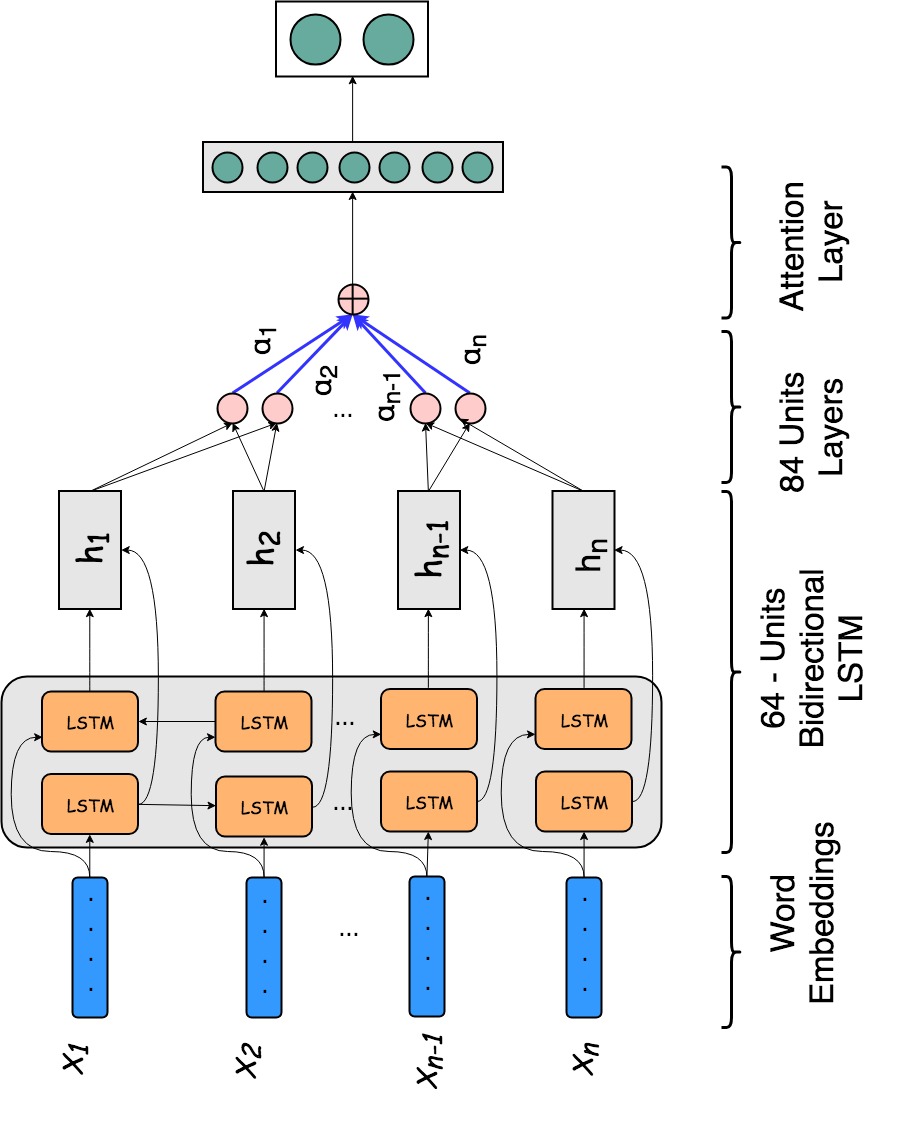}
  \caption{AttentionLSTM architecture consisting of one bidirectional LSTM layer, LeakyReLU activated and followed by an attention layer}
  \label{fig:atten}
\end{minipage}
\end{figure}

LSTM cell architectures tend to perform significantly better than regular neural networks in exploiting sequential data. An ordered input vector $\textbf{x}=(x_1, x_2,..,x_n)$ is mapped to an output vector $\textbf{y} = (y_1, y_2,...,y_n)$ throughout computing hidden state vector $\textbf{h} = (h_1, h_2,...,h_n)$ repetitively. Cell $i$ uses information from previous cells in time, creating a context feature, in the neural network architecture. However, sequential data, as text, speech or frame series often require knowledge of both past and future context. Bidirectional LSTMs solve this problem by exploiting input data, summarizing both the past and future context of each time sample. They assign two hidden states for the same time sample calculated in both directions in order to feed the output layer. The forward hidden sequence  $\overrightarrow{\textbf{h}}$ is calculated from reading input  $x_1$ to $x_n$ while on the other hand the backward hidden sequence  $\overleftarrow{\textbf{h}}$ reads input from $x_n$ to $x_1$. The total hidden state determines the time sample $i$ by concatenating both forward and backward hidden states, that is $h_i =\overrightarrow{h_i} ||\overleftarrow{h_i}$. In our work, we implement a deep two-layered bidirectional LSTM (BiLSTM) stacked with a dense layer between them as shown in Fig. \ref{fig:bilstm}.

\subsection{AttentionLSTM}
Additionally, a mechanism focusing on the most significant time sample is added on top of the regular LSTMs. This mechanism is called attention layer, was introduced by \cite{DBLP:journals/corr/BahdanauCB14} and maps hidden states according their importance. In other words, an attention layer assigns a value $\alpha_i$ to each state, which is called the attention factor, and is represented by a so-called attentive vector $r$:
\begin{equation}
   r_t = tanh(W_hh_t+b_t) 
\end{equation}
\begin{equation}
    a_t = sofmax(r_t) = \frac{e^{r_t}}{\sum_{j=0}^T {e^{r_t}}} ,\quad    \sum^n_{t=1}{a_t} = 1
\end{equation}
\begin{equation}
s = \sum_{t=0}^T {a_th_t} 
\end{equation}
where $W_h$ and $b_t$ are the LSTM model weights, optimized during training. As FL detection often demands focus on the sentiment contrast between words we implement an architecture based on an attentive LSTM layer (AttentionLSTM) as shown in Fig. \ref{fig:atten}. Finally, a dense softmax activation layer is applied to the $s$ feature vector representation of the tweet for the classification step. 

\subsection{Dense Neural Network}
\begin{figure}[!h]
\centering
\begin{minipage}[t]{.5\textwidth}
  \centering
        \includegraphics[width=1\textwidth]{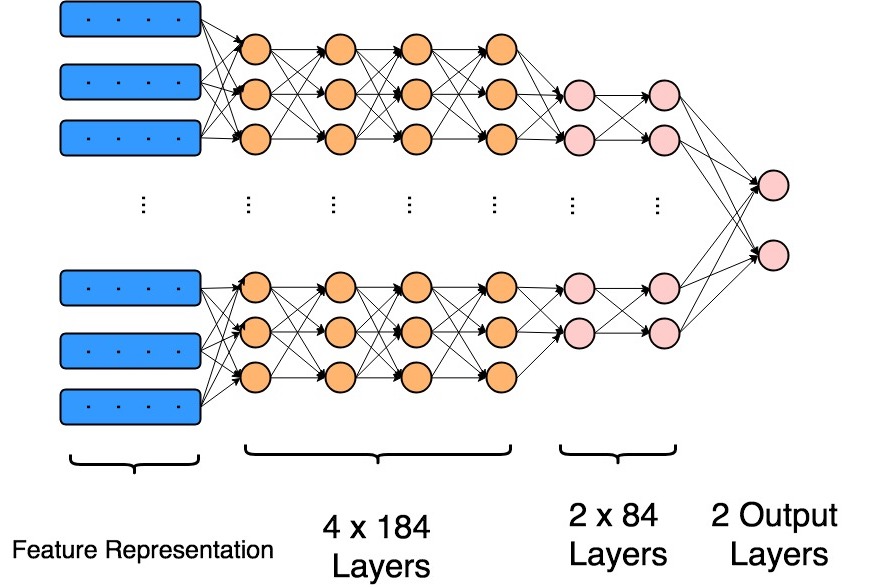}
  \caption{DNN  architecture}
  \label{fig:dense}
\end{minipage}%
\begin{minipage}[t]{.5\textwidth}
  \centering
      \includegraphics[width=1\textwidth]{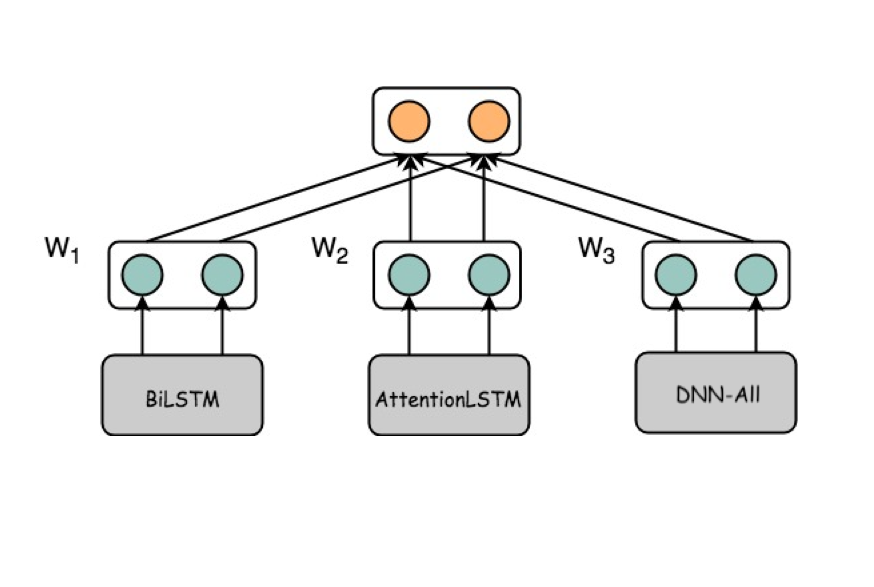}

  \caption{DESC: Deep Ensemble Soft Classifier architecture}
  \label{fig:ensemble}
\end{minipage}
\end{figure}

Finally, we implemented a dense fully connected deep neural network consisting of six layers. ReLU activation is used on every neuron and a dense vector representation of the features is presented the input as shown in Fig. \ref{fig:dense}. Detailed feature representations are presented in section \ref{sec:Features}.  
\subsection{Deep Ensemble Soft Classifier-DESC}
In order to capture FL in Twitter we use an ensemble technique, combining different classifiers. All three classifiers described above use disparate features at textual and word level. The ensemble method makes prediction based on weighted confidence scores from the three classifiers. The classifiers impact to the final classification is determined through cross validation during training phase. 
\begin{equation}
    w_i = \frac{e^{F_1^i}}{\sum_{i=1}^3{e^{F_1^i}}}, \quad i= 1, 2, 3 
\end{equation}
where $F_1^i$ stands for $F_1$ score of classifier $i$. The final prediction is made by combining the confidence scores of all three classifiers confidence scores multiplied with a scaling factor $w_i$. 

\begin{equation}
o_j = \operatorname*{arg\,max}_c  \sum_{i=1}^3 {w_i*\overrightarrow{p_i}^j},  \quad \overrightarrow{p_i}^j \in \mathbb{R}^c
\end{equation}
where j denotes the sample $j$ of data set, $\overrightarrow{p_i}^j$ the confidence score predicted by classifier $i$ for sample $j$ and c is the number of classes.

\section{Experimental Setup} 
\subsection{Datasets} 
 Detecting FL, in addition to its inherent difficulty as a problem, is a particularly difficult task since there is not much benchmark data available. To examine DESC robustness and reliability, we investigate three different Twitter-based datasets. First, we detect ironic comments on Twitter using SemEval-2018's 'Detecting Irony in English Twitter' balanced dataset \cite{hee_semeval-2018_2018}, consisting of 3834 train and
784, gold standard, test data. In addition, we utilized an imbalanced dataset, containing sarcastic tweets, compiled by Riloff et al \cite{riloff_sarcasm_2013}. Riloff's dataset is a high-quality dataset as indicated by its high Cohen's kappa score and consists of 2278 tweets, only 506 of them being sarcastic. Finally, to evaluate our model's performance on sentiment analysis we acquired a dataset containing all forms of figurative language as well as their sentiment polarity, as proposed on SemEval's 2015 Task-11 \cite{ghosh_semeval-2015_2015}. This dataset is also composed by figurative tweets and contains overall 8000 training and 4000 test tweets. Each tweet sample is ranked on a 11-point scale according to their sentiment polarity, ranging from –5 (negative sentiment polarity, for tweets with critical and ironic meanings) to +5 (positive sentiment polarity, for tweets with very upbeat meanings). 

\subsection{Feature Engineering} 
\label{sec:Features}
\begin{figure}[!h]
\centering
\begin{minipage}[t]{.5\textwidth}
  \centering
        \includegraphics[width=1.0\textwidth]{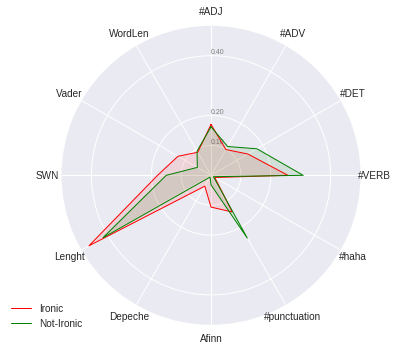}

\end{minipage}%
\begin{minipage}[t]{.5\textwidth}
  \centering
      \includegraphics[width=1.0\textwidth]{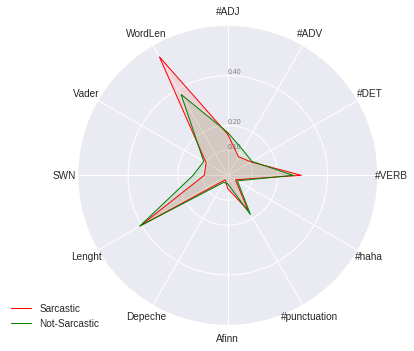}
    \newline

\end{minipage}

    \caption{Structural differences between irony and sarcasm}
    \label{fig:spider}
\end{figure}

As already mentioned, figurative language detection is a challenging NLP task, requiring complex attributes hopefully capturing non literal use of language. Every figure of speech tend to relate with linguistic patterns which discriminate it from other ones. By the same reasoning, we claim that if we could capture all aspects and structural differences between literal and figurative language we could enhance classification predictions. As demonstrated in Fig.\ref{fig:spider}, simple patterns such as sentiment polarity and average word length per tweet could denote non literal figures of speech.  Thus, we try to pursue the user's intention using syntactic, semantic and emotional language structure. We combine uni-gram and bi-gram models with 44 features extracted from text. These features are fed to the DNN and can be categorized into four major groups. 
\begin{itemize}
    \item \textbf{Syntactic Features (12)}: This group contains features indicative of the user's syntactic usage habits, and consists of the frequency of the Part-Of-Speech(POS) tags.
    \item \textbf{Demonstrative Features (8)}: Attributes implying user's expression, such as: the number of words and emojis used, average length of words in the tweet, frequency of punctuation marks, duplicate letters that may express user's emphasis, as well as the frequency of polysyllabic words.
    \item \textbf{Sentiment Features (12)}: As argued before, FL can be defined as sentiment contrast between words in short distance. This necessitated the use of many lexicons, in order to obtain a sentiment estimation in word level, and then estimate the overall tweet positive, negative and contradictory sentiment, as shown in Eq.\ref{contrast}. For this purpose, we calculate the tweet's average positive and negative sentiment using SentiWordNet\cite{baccianella_sentiwordnet_2010}, VADER\cite{hutto_vader:_2014}, Afinn \cite{aarup_nielsen_new_2011}, DepecheMood \cite{staiano_depechemood:_2014}, as well as their positive-negative variance by the following formulas: 

    \begin{equation}
        \label{contrast}
        S^{pos}_{i} = \frac{1}{n_i}\sum_{j=1}^{n_i}w^{pos}_{i,j},\quad
        S^{neg}_{i} = \frac{1}{n_i}\sum_{j=1}^{n_i}w^{neg}_{i,j},\quad
        S^{contrast}_{i} = S^{pos}_{i} - S^{neg}_{i}
    \end{equation}

    \item \textbf{Mood Features (8)}: An additional lexicon, DepecheMood, was used to annotate words indicating the user's temper. Mood features describe user's happiness, sadness, annoyance, inspiration, fear, indifference, anger and amusement and can help us to detect non literal quotes. 
    \item \textbf{Readability Features (4)}: Finally, we claim that all FL forms tend to have different readability scores than literally ones. Thus, we implement four metrics measuring text readability using well known readability scores. First, we enumerate words not present on Dale Chall's list and afterwards we calculate three readability scores:
       \begin{flalign}
            \textrm{Dale Chall} = \displaystyle 0.1579\left({\frac {\mbox{difficult words}}{\mbox{words}}}\times 100\right)+0.0496\left({\frac {\mbox{words}}{\mbox{sentences}}}\right) &&
        \end{flalign}
        \begin{flalign}
            \textrm{Flesch} = \displaystyle 206.835-1.015\left({\frac {\textrm{total words}}{\textrm{total sentences}}}\right)-84.6\left({\frac {\textrm{total syllables}}{\textrm{total words}}}\right) &&
        \end{flalign}
        \begin{flalign}
            \textrm{Gunning Fog} =\displaystyle 0.4\left[\left({\frac {\mbox{words}}{\mbox{sentences}}}\right)+100\left({\frac {\mbox{complex words}}{\mbox{words}}}\right)\right] &&
        \end{flalign}
    
\end{itemize}

\section{Experimental Results}
We compared the proposed DESC method against a variety of classifiers and different feature combinations. We experimented with different combinations of Tf-Idf and the features presented in section \ref{sec:Features} using a deep neural network architecture and also an SVM, frequently used by the teams participating in SemEval. As illustrated in Table \ref{eswteriko}, DESC outperforms all baseline classifiers with consistent (over all metrics) performance figures, a fact that is indicative for the robustness of the proposed approach. In addition, we can observe the stability that DESC demonstrates on all datasets, including unbalanced and multiclass-classification problems. The abbreviations used in the table for the features are unigrams, bigrams along with feature Feature Vector (FeatVec) (2inp), unigrams (uni) only , Tf-idf (Tfidf), the feature set of Section \ref{sec:Features} (FeatVec) as well the concatenation of all the features (All).
\begin{table}[]
\caption{Comparison with baseline classifiers (various set-ups of DNN, SVM, AttLSTM, and BiLSTM) for the tasks of (a) irony and sarcasm detection (binary), and (b) sentiment polarity detection (eleven ordered values, -5 for "very negative" to 5 for"very positive") - bold figures indicate superior performance. \\}
\centering
\scalebox{0.74}{
\begin{tabular}{l|c|c|c|c|c|c|c|c|c|c|c|c|c|c|c|c|c|}
\cline{2-18}
                                    & \multicolumn{15}{l|}{(a) Irony \& Sarcasm detection}                                                                                                                                                                                          & \multicolumn{2}{l|}{\begin{tabular}[c]{@{}l@{}}(b) Sentiment\\  polarity\\  detection\end{tabular}} \\ \cline{2-18} 
                                    & \multicolumn{5}{l|}{\begin{tabular}[c]{@{}l@{}}Irony\\ SemVal-2018-Task 3.A\cite{hee_semeval-2018_2018}\end{tabular}}                              & \multicolumn{5}{l|}{\begin{tabular}[c]{@{}l@{}}Sarcasm\\ Riloff \cite{riloff_sarcasm_2013}\end{tabular}}                                         & \multicolumn{5}{l|}{Average}                                                  & \multicolumn{2}{l|}{\begin{tabular}[c]{@{}l@{}}Sentiment/\\ SemVal2015\\ Task 11\cite{ghosh_semeval-2015_2015}\end{tabular}}      \\ \hline
\multicolumn{1}{|l|}{System}        & Acc           & Pre           & Rec           & F1            & AUC           & Acc           & Pre           & Rec           & F1            & AUC           & Acc           & Pre           & Rec           & F1            & AUC           & COS                                              & MSE                                              \\ \hline
\multicolumn{1}{|l|}{DNN-2inp}      & 0,63          & 0,64          & 0,62          & 0,63          & 0,65          & 0,82          & 0,78          & 0,81          & 0,79          & 0,84          & 0,73          & 0,71          & 0,72          & 0,71          & 0,75          & 0,602                                            & 4,230                                            \\ \hline
\multicolumn{1}{|l|}{DNN-Tfidf}     & 0,65          & 0,65          & 0,63          & 0,64          & 0,70          & 0,83          & 0,81          & 0,83          & 0,80          & 0,72          & 0,74          & 0,73          & 0,73          & 0,72          & 0,71          & 0,710                                            & 3,170                                            \\ \hline
\multicolumn{1}{|l|}{DNN-uni}       & 0,65          & 0,68          & 0,65          & 0,65          & 0,72          & 0,79          & 0,78          & 0,81          & 0,79          & 0,74          & 0,72          & 0,73          & 0,73          & 0,72          & 0,73          & 0,690                                            & 8,430                                            \\ \hline
\multicolumn{1}{|l|}{DNN-All}       & 0,66          & 0,69          & 0,66          & 0,67          & 0,75          & 0,83          & 0,81          & 0,83          & 0,81          & 0,82          & 0,75          & 0,75          & 0,75          & 0,74          & 0,79          & 0,789                                            & 2,790                                            \\ \hline
\multicolumn{1}{|l|}{DNN-FeatVec}   & 0.64          & 0.65          & 0.65          & 0.65          & 0.71          & 0,81          & 0,81          & 0,83          & 0,80          & 0,72          & 0,73          & 0,73          & 0,74          & 0,73          & 0,71          & 0,680                                            & 3,230                                            \\ \hline
\multicolumn{1}{|l|}{SVM-Tfidf}     & 0,65          & 0,68          & 0,65          & 0,66          & 0,70          & 0,82          & 0,80          & 0,82          & 0,80          & 0,80          & 0,74          & 0,74          & 0,74          & 0,73          & 0,75          & 0,720                                            & 2,890                                            \\ \hline
\multicolumn{1}{|l|}{SVM-FeatVec}   & 0,59          & 0,59          & 0,59          & 0,59          & 0,60          & 0,82          & 0,73          & 0,81          & 0,75          & 0,76          & 0,71          & 0,66          & 0,70          & 0,67          & 0,68          & 0,700                                            & 3,390                                            \\ \hline
\multicolumn{1}{|l|}{SVM-All}       & 0,66          & 0,69          & 0,66          & 0,67          & 0,75          & 0,83          & 0,81          & 0,83          & 0,81          & 0,81          & 0,75          & 0,75          & 0,75          & 0,74          & 0,78          & 0,723                                            & 2,810                                            \\ \hline
\multicolumn{1}{|l|}{AttentionLSTM} & 0,71          & 0,70          & 0,71          & 0,70          & 0,75          & 0,85          & 0,83          & 0,85          & 0,83          & 0,84          & 0,78          & 0,77          & 0,78          & 0,77          & 0,80          & 0,749                                            & 2,860                                            \\ \hline
\multicolumn{1}{|l|}{BiLSTM}        & 0,71          & 0,71          & 0,71          & 0,70          & 0,76          & 0,85          & 0,85          & 0,85          & 0,85          & 0,85          & 0,78          & 0,78          & 0,78          & 0,78          & 0,81          & 0,704                                            & 3,220                                            \\ \hline
\multicolumn{1}{|l|}{\textbf{DESC}} & \textbf{0,74} & \textbf{0,73} & \textbf{0,73} & \textbf{0,73} & \textbf{0,78} & \textbf{0,87} & \textbf{0,87} & \textbf{0,86} & \textbf{0,87} & \textbf{0,86} & \textbf{0,81} & \textbf{0,80} & \textbf{0,80} & \textbf{0,80} & \textbf{0,82} & \textbf{0,820}                                   & \textbf{2,480}                                   \\ \hline
\end{tabular}
}

\label{eswteriko}
\end{table}

We tested DESC against the performance scores of all models submitted and published in SemEval-2015 \cite{ghosh_semeval-2015_2015} Sentiment Analysis task.
DESC achieves 0.82 in cosine similarity whereas the winning team \cite{ozdemir_clac-sentipipe:_2015} obtained 0.758; ranked also on 4th position regarding the MSE measure. At the same time, ClaC and UPF teams are the only one to obtain a better MSE value than DESC, as illustrated in Table \ref{tab:sent}. Further evidence of the robustness of DESC across all metrics on both Irony \cite{hee_semeval-2018_2018} and Sarcasm \cite{riloff_sarcasm_2013} detection is illustrated in Table \ref{tab:irony} and Table \ref{tab:rilof}. Specifically, DESC outperforms all submissions on \cite{hee_semeval-2018_2018} regarding F1 measure; a fact that it is indicative for the balance achieved between precision and recall figures which also satisfies the desired property for low false-positives regrading the task of automated twitter classification. In addition, DESC's performance is highly increased compared to Riloff's initial proposal \cite{riloff_sarcasm_2013} and is very close to Ghosh\&Veale \cite{ghosh_fracking_2016}. 

\begin{table}[!ht]
\begin{minipage}[t]{0.5\linewidth}
\centering
\caption{Systems comparison on SemEval-2018-Task 3-$A^*$ ironic dataset.
\small{*The five best performing models, as officially reported in \cite{hee_semeval-2018_2018}}}
\begin{tabular}[b]{|l||c|c|c|c|}
\hline

Submission      & Acc           & Pre           & Rec           & $F_1$            \\ \hline
THU\_NGN        & 0,73          & 0,63          & 0,80          & 0,71          \\ \hline
NTUA-SLP        & 0,73          & 0,65          & 0,69          & 0,67          \\ \hline
WLV             & 0,64          & 0,53          & \textbf{0,84} & 0,65          \\ \hline
rangwani\_harsh & 0,66          & 0,55          & 0,79          & 0,65          \\ \hline
NIHRIO, NCL     & 0,70          & 0,61          & 0,69          & 0,65          \\ \hline
\textbf{DESC}   & \textbf{0,74} & \textbf{0,73} & 0,73          & \textbf{0,73} \\ \hline
\end{tabular} 

\label{tab:irony}
\end{minipage}
\hspace{0.5cm}
\begin{minipage}[t]{.5\linewidth}
\centering
\caption{Systems comparison on Riloff's imbalanced sarcastic dataset.\\}
\begin{tabular}[b]{|l|c|c|c|}
\hline

System submission       & Pre        & Rec       & $F_{1}$        \\ \hline
Riloff                  & 0,44       & 0,62      & 0,51      \\ \hline
Ghosh\&Veale            & \textbf{0,88}       & \textbf{0,88}      & \textbf{0,88}      \\ \hline
DESC                    & 0,87       & 0,86      & 0,87      \\ \hline
\end{tabular}
\label{tab:rilof}
\end{minipage}
\end{table}

\begin{figure}[!h]
  \centering
      \includegraphics[width=0.7\textwidth   ]{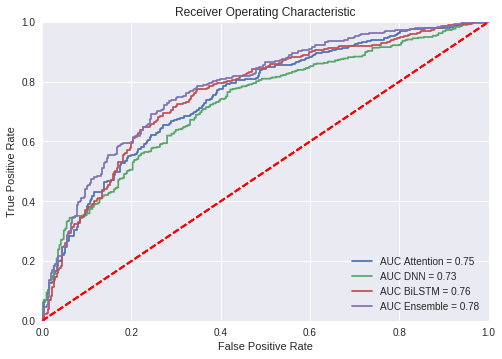}
  \caption{ROC-AUC curve for three ensemble classifiers on SemEval's-2018 Irony detection task. }
  \label{fig:auc}
\end{figure}

\begin{table}[!ht]
\centering
\caption{Systems comparison on Sentiment analysis SemEval-2015-Task $11^{*}$  according to task's official metrics, Mean Squarred Error and Cosine similarity. \small{*Officialy submitted results reported in \cite{ghosh_semeval-2015_2015}. From the total of 15 submissions only the ones that exhibit higher performance than the average over both metrics are included. The first three entries refer to reported values for three baseline classifiers.}\\}
\begin{tabular}{|l|c|c|}
\hline
System submission & Cosine        & MSE            \\ \hline
Naive Bayes      & 0,39          & 5,672          \\ \hline
MaxEnt            & 0,43          & 5,450          \\ \hline
Decision Tree     & 0,55          & 4,065          \\ \hline
CLaC              & 0,76          & \textbf{2,117} \\ \hline
UPF               & 0,71          & 2,458          \\ \hline
LLT\_PolyU        & 0,69          & 2,600          \\ \hline
LT3               & 0,66          & 2,913          \\ \hline
ValenTo           & 0,63          & 2,999          \\ \hline
PRHLT             & 0,62          & 3,023          \\ \hline
CPH               & 0,62          & 3,078          \\ \hline
elirf             & 0,66          & 3,096          \\ \hline
\textbf{DESC}     & \textbf{0,82} & 2,480          \\ \hline
\end{tabular}

\label{tab:sent}

\end{table}
Finally, combining all three AttentionLSTM, BiLSTM and DNN-all classifiers in an Ensemble model we can detect Irony with increased confidence as illustrated in Fig \ref{fig:auc}.

\section{Conclusion \& Future Work} 
In this paper, we propose an ensemble method for sentiment classification and Figurative Language tracking with  non contextual information, in short texts segments like tweets. The proposed method is based on a combination of word and sentence based features and outperforms published SemEval models in most metrics and on all datasets. In addition, we introduce a feature level approach using a variety of author's sentiment, mood and expressiveness to detect FL usage. Beyond that, DESC shows improved AUC performance at the difficult task of irony detection.  

DESC model could be even more accurate if we enhance past-present diversion since sarcastic utterances tend to include more time-related contradictions as Rajadesingan \cite{rajadesingan_sarcasm_2015} denoted. Beside time and sentiment contradictions a behavioral approach could be used to analyze change's in user's stylistic patterns such as average tweet length, user's word style familiarity or even user's sarcastic intention.  Another key improvement could be obtained by measuring the gap between written and informal spoken style using the ANC Frequency Data corpus as Barbieri \cite{barbieri_modelling_2014} proposed. In addition, Linguistic Inquiry and Word Count (LIWC) could be used to extract vectorized lingual patterns indicating FL. Finally, the approach presented in this paper is based solely on data from Twitter's short texts. A future FL system could become even more effective if trained and tuned using texts including FL from more sources.  
\footnote{Work originally published in Engineering Applications of Neural Networks (EANN)-2019 \cite{potamias2019robust}. An extension of this paper can be found in \cite{potamias2020transformer}.} 

%
%
\bibliographystyle{splncs04}
\bibliography{Thesis.bib}

\end{document}